\documentclass[pdflatex,sn-mathphys]{sn-jnl}

\definecolor{green}{rgb}{0, 0.5, 0}
\definecolor{orange}{rgb}{0.8, 0.6, 0.2}
\definecolor{red}{rgb}{1.0, 0.0, 0.0}
\definecolor{teal}{rgb}{0.0, 0.4, 0.4}
\definecolor{purple}{rgb}{0.65,0,0.65}
\definecolor{saffron}{rgb}{0.95,0.75,0.2}
\definecolor{turquoise}{rgb}{0.0,0.5,0.5}
\definecolor{black}{rgb}{0.0, 0.0, 0.0}
\definecolor{gray}{rgb}{0.5, 0.5, 0.5}

\newcommand{\rui}[1]{{\color{black}#1}}

\newcommand{\wy}[1]{{\color{black}#1}}

\newcommand{\rv}[1]{{\color{black}#1}}

\jyear{2022}%

\theoremstyle{thmstyleone}%
\theoremstyle{thmstyletwo}%

\theoremstyle{thmstylethree}%

\begin{document}

\title[Shape-Aware Fine-Grained Classification of Erythroid Cells]{Shape-Aware Fine-Grained Classification of Erythroid Cells}


\author[1]{\fnm{Ye} \sur{Wang}}\email{wangyejlu1996@gmail.com}
\equalcont{These authors contributed equally to this work.}
\author[2,4]{\fnm{Rui} \sur{Ma}}\email{ruim@jlu.edu.cn}
\equalcont{These authors contributed equally to this work.}

\author[1]{\fnm{Xiaoqing} \sur{Ma}}\email{maxq20@mails.jlu.edu.cn}
\author[3]{\fnm{Honghua} \sur{Cui}}\email{chh1214@126.com}
\author[1]{\fnm{Yubin} \sur{Xiao}}\email{xiaoyb@smail.xtu.edu.cn}
\author[1]{\fnm{Xuan} \sur{Wu}}\email{wuuu20@mails.jlu.edu.cn}
\author*[1]{\fnm{You} \sur{Zhou}}\email{zyou@jlu.edu.cn}

\affil*[1]{\orgdiv{School of Computer Science and Technology}, \orgname{Jilin University}, \orgaddress{\city{Changchun}, \postcode{130012}, \state{Jilin Province}, \country{China}}}

\affil[2]{\orgdiv{School of Artificial Intelligence}, \orgname{Jilin University}, \orgaddress{\city{Changchun}, \postcode{130012}, \state{Jilin Province}, \country{China}}}

\affil[3]{\orgdiv{Second Hospital of Jilin University}, \orgaddress{ \city{Changchun}, \postcode{130041}, \state{Jilin Province}, \country{China}}}

\affil[4]{\orgdiv{Engineering Research Center of Knowledge-Driven Human-Machine Intelligence}, \orgname{MOE}, \orgaddress{ \city{Changchun}, \postcode{130012}, \state{Jilin Province}, \country{China}}}



\abstract{
Fine-grained classification and counting of bone marrow erythroid cells are vital for evaluating the health status and formulating therapeutic schedules for leukemia or hematopathy.
Due to the subtle visual differences between different types of erythroid cells, it is challenging to apply existing image-based deep learning models for fine-grained erythroid cell classification. 
Moreover, there is no large open-source datasets on erythroid cells to support the model training.
In this paper, we introduce BMEC (Bone Morrow Erythroid Cells), the first large fine-grained image dataset of erythroid cells, to facilitate more deep learning research on erythroid cells.
BMEC contains 5,666 images of individual erythroid cells, each of which is extracted from the bone marrow erythroid cell smears and professionally annotated to one of the four types of erythroid cells.
To distinguish the erythroid cells, one key indicator is the cell \textit{shape} which is closely related to the cell growth and maturation.
Therefore, we design a novel shape-aware image classification network for fine-grained erythroid cell classification. 
The shape feature is extracted from the shape mask image and aggregated to the raw image feature with a shape attention module.
With the shape-attended image feature, our network achieved superior classification performance (81.12\% top-1 accuracy) on the BMEC dataset comparing to the baseline methods.
\rv{Ablation studies also demonstrate the effectiveness of incorporating the shape information for the fine-grained cell classification.}
To further verify the generalizability of our method, we tested our network on two additional public white blood cells (WBC) datasets and the results show our shape-aware method can generally outperform recent state-of-the-art works on classifying the WBC.
The code and BMEC dataset can be found on \url{https://github.com/wangye8899/BMEC}.
}


\maketitle

\section{Introduction}
The erythrocytes or red blood cells are one of the most critical cells in the body. There are four types of erythroid cells (Figure \ref{fig:cell_type}) based on their growth and maturation.
In a healthy body, they maintain a relatively balanced state. However, in many blood disorders such as anemia, the number of these erythroid cells will change and become unbalanced. 
Therefore, the \textit{fine-grained} classification and counting of different types of erythroid cells are vital in diagnosing and preventing the related diseases \cite{chitra2019detection}.

\rui{One typical way} for cell classification and counting is to manually observe the morphological or shape differences of the cells through the microscope. 
Such process relies heavily on the hematologists' experience and skills, and it is easy to make mistakes and introduce subjectivity to the results.
Furthermore, it may take several years to train a competent hematologist to distinguish the specific type of erythroid cells.
In recent years, several image processing and machine learning techniques for classifying and counting erythroid cells have been proposed.
Most existing methods \cite{alomari2014automatic,lippeveld2020classification,petrovic2020sickle} mainly \rui{follow the traditional machine learning pipeline}, including data prep-processing, feature extraction, feature selection, and classification steps.
Since these methods rely on manually designed features, the generalizability of the classification model may be limited.
In addition, the morphological and shape differences are not fully and deeply exploited in the model.
Even though satisfied results have been achieved for classifying white blood cells or certain abnormal types of erythroid cells whose shapes are visibly distinctive, traditional methods may not perform well for fine-grained classification of erythroid cells with only subtle shape differences.

On the other hand, deep learning techniques have shown superior performance than traditional machine learning methods in computer vision tasks such as image classification \cite{liu2021swin}, object detection \cite{xu2021end} and segmentation \cite{strudel2021}.
Deep neural networks can automatically extract deep image features which are more robust and more generalizable than hand-crafted features. 
However, most pre-trained network models \cite{simonyan2014very, he2016deep, huang2017densely, HangZhang2020ResNeStSN, liu2021swin} cannot directly learn the subtle shape differences which are crucial for fine-grained classification of erythroid cells.
To enable fine-grained erythroid cell classification, a more customized and shape-aware network needs to be designed and trained on erythroid cell images with finer-level category annotation.
Unfortunately, unlike the leukocytes or white blood cells (WBC), for which many open-source datasets are available (e.g., LISC \cite{rezatofighi2011automatic}, BCCD \cite{mohamed2012efficient} and Raabin-WBC \cite{kouzehkanan2021raabin}), \rui{there is no existing open-source fine-grained datasets to support the deep learning on erythroid cells}.


In this paper, we introduce BMEC (Bone Morrow Erythroid Cells), a large fine-grained image dataset of erythroid cells, to facilitate the study for erythroid cells in real clinical scenarios.
BMEC dataset contains 5,666 images of individual erythroid cells.
Each image contains one erythroid cell extracted from the bone marrow erythroid cell smears and is annotated to one of the four types of erythroid cells by the hematologists (Figure \ref{fig:cell_type}).
To classify the erythroid cells into fine-grained categories, we propose a novel \textit{shape-aware} image classification model which explicitly encode the shape information into the network (Figure \ref{fig:overview}).
Specifically, we first extract the cell shape mask image from the input erythroid cell image.
Then, the shape mask feature and the input image feature are extracted using the existing backbone image feature extraction models.
A shape attention module is further employed to compute a shape-attended image feature, which is aggregated to the shape feature and the raw image feature to form a fused feature for final classification.
 
We conducted extensive quantitative evaluations of different backbone models for our shape-aware network on BMEC and our final model with Swin-Transformer \cite{liu2021swin} and VGG19 \cite{simonyan2014very} achieved 81.12\% top-1 accuracy which is consistently superior than other backbone model combinations.
We also performed ablation studies to verify the effectiveness of explicitly using shape information for fine-grained erythroid cell classification.
Furthermore, we tested our shape-aware classification network on two public leukocytes or WBC datasets and the results show our method can outperform recent state-of-the-art work on classifying the WBC.
In summary, our contributions are as follows:

\begin{itemize}
\item We introduce BMEC, the first large dataset of erythroid cell images with fine-grained category annotation, to support more deep learning research on erythroid cells. This dataset contains 5,666 professionally annotated images of four types of erythroid cells, while there are only subtle shape differences between these fine-grained types, making our dataset more chanllenging than existing WBC datasets.


\item We design a novel shape-aware network which employs a shape attention module to compute a shape-attended image feature and uses the shape-aggregated feature to improve the fine-grained classification accuracy of erythroid cells.
To our best knowledge, we are the first to explicitly incorporate the shape information to the deep learning framework for this task.


\item We perform extensive evaluations of our network with different backbone models and conduct ablation studies to verify the effectiveness of shape information.
The results on BMEC and two other WBC datasets show the superiority of the proposed shape-aware network and models in classifying the blood cell images.
\end{itemize} 

\section{Related Work}

\begin{table}[t]

\begin{center}
\begin{minipage}{350pt}
\caption{Comparison of related WBC and erythroid cell datasets with our BMEC.}\label{tab:datasets}%
\centering
\setlength{\tabcolsep}{2pt}
\begin{tabular}{@{}lclcll@{}}
\toprule
Dataset& Quantity& Cell Type& Classes& Label& Microscope\\
\midrule
LISC \cite{rezatofighi2011automatic}& 250& WBC& 5& one expert& Axioskope40\\  
BCCD \cite{mohamed2012efficient}    & 349& WBC& 5& one expert    & Regular light micro\\
Raabin-WBC\cite{kouzehkanan2021raabin}& 17,965& WBC& 5& two experts& Olympus Cx18\\     
CBC \cite{alam2019machine} & 360& Erythroid cells& 3& two experts& ECLIPSE 50i micro\\
erythrocytesIDB \cite{gonzalez2014red} & 629& Erythroid cells & 3& one expert& Leika micro\\
BMEC (ours) & 5,666& Erythroid cells & 4& three experts & Olympus BX43 \\
\botrule
\end{tabular}
\end{minipage}
\end{center}
\end{table}


In this section, we first discuss several datasets on blood cells, including both leukocytes and erythrocytes, with the emphasis on the data characteristics and how they are collected; see Table \ref{tab:datasets}. 
We then provide a brief review of the traditional and deep learning techniques related to blood cell classification and counting.

\subsection{Blood Cell Datasets}
\textbf{WBC datasets.} There are many available white blood cell image datasets for investigating the WBC classification and counting tasks \cite{rezatofighi2011automatic,mohamed2012efficient,labati2011all,zheng2018fast,naruenatthanaset2020red,kouzehkanan2021raabin}.
Here, we only describe three representative WBC datasets and highlight their difference with our BMEC; see Table \ref{tab:datasets}.
The LISC \cite{rezatofighi2011automatic} is an early WBC dataset that includes the hematological images taken from peripheral blood of healthy subjects.
The smears were stained by through the Gismo-right technique and a Sony Model No.SSCDC50AP camera was used to capture the observation of a Axioskope 40 microscope at 100X magnification. 
Then, 250 WBC images were extracted and labelled by one blood expert into five fine-grained categories.
%
Similarly, BCCD \cite{mohamed2012efficient} contains 349 WBC images that have been taken from peripheral blood and annotated by one expert.
The smears were also stained by the Gismo-right technique and the observations of a regular light microscope at 100X were captured using a CCD color camera.
Each WBC image extracted from the smear image was annotated by one expert into one of five categories.
Raabin-WBC \cite{kouzehkanan2021raabin} is a recent, large and public dataset of white blood cell images extracted from the normal peripheral blood samples. 
The peripheral blood smears were stained by the Giemsa technique and the images were taken using a Samsung Galaxy S5 camera \rui{to capture the observation of an Olympus CX18 microscope at 100X magnification}. 
\wy{In total, 17,965 WBCs images are cropped from the smear images and labelled into five categories by two blood experts. }
Comparing to the white blood cells, different types of erythroid cells look more similar and the shape difference between the types is subtle.
Therefore, it is more challenging for fine-grained erythroid cell classification, e.g., over 99\% accuracy can be achieved on Raabin-WBC while only around 80\% accuracy can be obtained on our BMEC.
Due to the difficulty on distinguishing the erythroid cell types, we asked three hematologists to jointly label the BMEC images in a more rigorous manner.

\textbf{Erythroid cell datasets.} Compared to the WBC datasets, there are only a few  open-source datasets containing the erythrocytes, while the data quality, quantity and variety are limited comparing to our BMEC.
As shown in Table \ref{tab:datasets}, CBC \cite{alam2019machine} is a hybrid dataset which contains three types of blood cells: red blood cells, white blood cells, and platelets.
\wy{The images in CBC were taken using a Nikon V1 camera to capture the observation of an ECLIPSE 50i microscope at 100X magnification, and then annotated by two experts.}
It focuses more on investigating the differences between the three types of blood cells rather than classifying red blood cells into fine-grained types. In addition, CBC only has 360 images which are difficult for training the deep neural networks.
One most related dataset on erythroid cell classificaiton is the erythrocytesIDB \cite{gonzalez2014red} which contains 629 images of individual erythroid cells extracted from 196 full filed peripheral blood images. 
A Kodak EasyShare V803 camera was used to capture the observation of a Leika microscope at 100X magnification.
The erythrocytesIDB images are annotated by \wy{one} expert into three categories based on the cell shape: circular, elongated and other shape.
Comparing to BMEC, the shape differences for erythroid cells in erythrocytesIDB are more significant since they focus on the study in cell deformation caused by certain blood diseases.
Hence, it's much easier to achieve high accuracy (98\% in \cite{gonzalez2014red}) on erythrocytesIDB.
In contrast, our BMEC provides a much larger number of images with minor shape differences, so that it can be used to train a more generalizable deep learning model for fine-grained erythroid cell classificaiton.


\subsection{Cell Classification Techniques}
Along with the release of more open-source blood cell datasets, automatic cell classification and counting techniques have been proposed, including the traditional image processing and machine learning based methods, as well as the deep learning based frameworks.

\textbf{Traditional methods.} 
Generally, cell classification using traditional image processing and machine learning techniques involves image segmentation, feature extraction, classification, and counting steps.
For example, Alomari et al. \cite{alomari2014automatic} employed the Circular Hough Transform (CHT) in an iterative circle detection framework to segment, classify and count WBCs and RBCs.
Such hough transform based shape detection methods are usually time-consuming and their detection and classification accuracy may be significantly reduced when the shape of the blood cells becomes more complex.
Meanwhile, machine learning based methods aim to classify the blood cells based on some features (e.g., shape and color) extracted from the cell images.
Lippeveld et al. \cite{lippeveld2020classification} applied the traditional machine learning techniques such as random forest \cite{breiman2001random} and gradient boosting classifier \cite{friedman2002stochastic} to classify and count the stain-free WBCs images \rui{based on the size, location, texture, and signal strength feature}.
Tavakoli et al. \cite{tavakoli2021new} proposed a new segmentation and feature extraction algorithm which computes three shape features and four novel color features for classifying WBC images.
Although above traditional methods have achieved appealing results for WBC or RBC classification on certain datasets such as Raabin-WBC and LISC, they heavily rely on hand-crafted features which are hardly generalizable to more challenging tasks, e.g., fine-grained classification on our BMEC.   

\textbf{Deep learning based cell classification.}
Deep learning can automatically extract features by learning from a large amount of data to compute tasks such as \rv{classification \cite{liu2021subtler,wang2022multilayer} and detection \cite{sun2021mask,zhou2022erythroidcounter}}.
To apply deep learning models for the cell classification, one way is to use models pre-trained on general large image datasets for feature extraction and perform particular traditional feature selection and classification methods on the smaller-scale WBC or RBC dataset.
In \cite{tougaccar2020classification}, features extracted from pre-trained AlexNet \cite{krizhevsky2012imagenet}, GoogLeNet \cite{szegedy2015going} and ResNet50 \cite{he2016deep} models are filtered using the Maximal Information Coefficient and Ridge feature selection methods, and quadratic discriminant analysis was used as a classifier on a dataset of WBCs.
Sahlol et al. \cite{sahlol2020efficient} employed a statistically enhanced Salp Swarm Algorithm\cite{mirjalili2017salp} to filter the features extracted from WBC images using the VGG \cite{simonyan2014very} model pre-trained on ImageNet and performed the classification using a traditional decision tree.

The other way to apply deep learning for cell classification is using the transfer learning technique which starts from a pre-trained CNN model and performs model fine-tuning on a specific WBC or RBC dataset.
For example, Alzubaidi et al. \cite{alzubaidi2020deep} first trained a CNN model on a combination of three WBC datasets and then employed transfer learning to classify RBC images in erythrocytesIDB \cite{gonzalez2014red}.
\wy{
Pasupa et al. \cite{pasupa2020convolutional} exploited pre-trained ResNet \cite{he2016deep} and DenseNet \cite{huang2017densely} and performed fine-tuning on their RBC images. In addition, they have handled the class imbalance problem by incorporating the focal loss \cite{lin2017focal}
}.
Comparing to traditional machine learning techniques, deep learning based classification methods can achieve superior performance without manual feature design. 
However, it is difficult to either train a traditional classifier or perform transfer learning for the fine-grained classification task, especially when there is no sufficient finely annotated training data.
We propose a novel shape-aware transfer learning framework that utilizes different types of pre-trained backbone models and fine-tunes the network on our large-scale BMEC for fine-grained erythroid cell classification.

\textbf{Shape-aware cell classification.}
The cell's morphological or shape characteristics such as the circularity, convexity and size have been investigated in traditional methods \cite{alomari2014automatic, lippeveld2020classification, tavakoli2021new} for cell classification.
For deep learning based methods, given only the original cell images, the networks can automatically learn some shape features \cite{zeiler2014visualizing} which are generally more robust than the hand-crafted ones.
However, the implicit learning of deep shape features requires a large amount of training data.
Even when transfer learning is used, it may be still difficult for the network to learn the fine-grained shape variations.
To explicitly incorporate the shape information in the network, Tavakoli et al. \cite{tavakoli2021generalizability} passed two binary shape images along with the RGB images of WBCs to a customized CNN with five channels (three for RGB and two for shape masks) and showed the generalizability of the model can be improved when the shape masks are used.
Similarly, we also explicitly integrate the shape information into the network.
Instead of using the nucleus and ROC (Representative Of
the Convex hull = convex hull - nucleus) images as in \cite{tavakoli2021generalizability}, we choose convex hull as our shape mask image since we find the contour is more suitable to represent the shape of erythroid cells.
Moreover, we design a novel shape attention module which computes a shape-attended image feature, to improve the learning of fine-grained shape-aware image features.

\begin{figure*}[!t]
    \centering
    \includegraphics[width=0.95\linewidth]{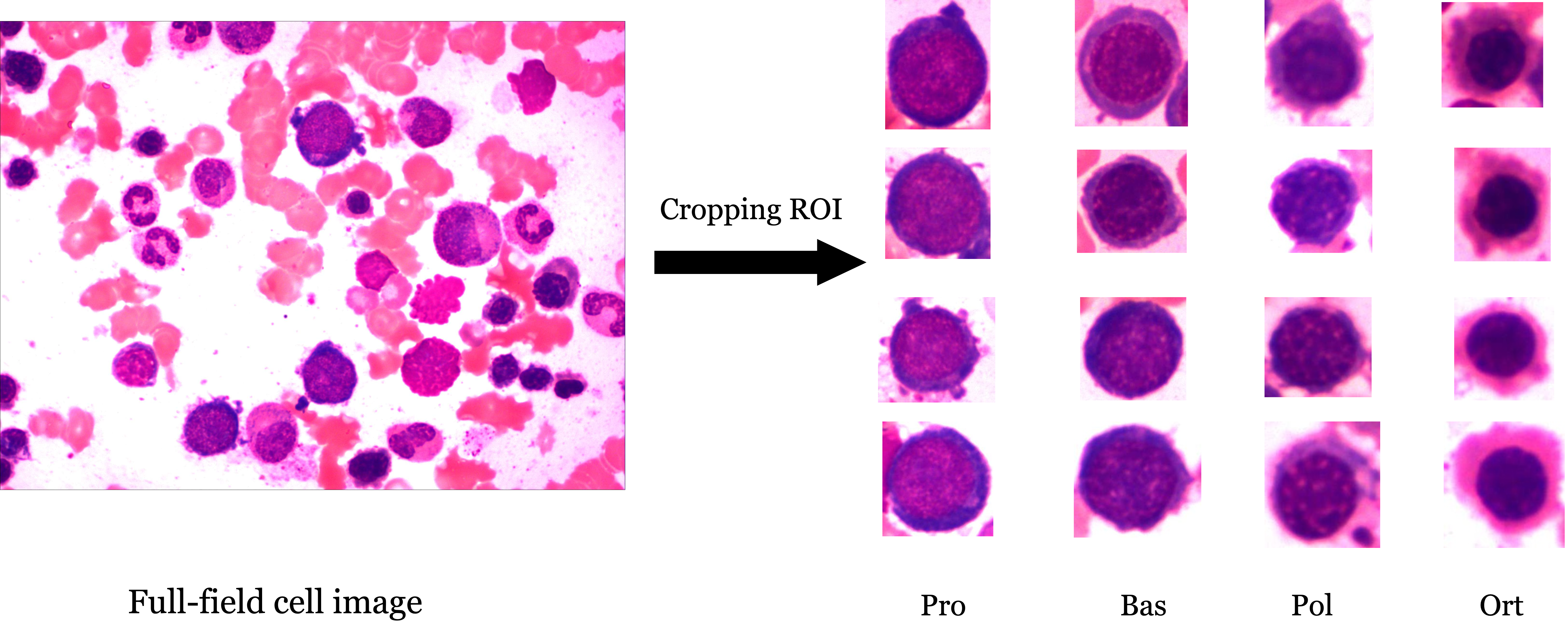}
    \caption{ Full-field erythroid cell image and  four different types of erythroid cells in BMEC cropped by manually annotated ROI: proerythroblast (Pro), basophilic erythroblast (Bas), polychromatophilic erythroblast (Pol), and Orthochromic erythroblast (Ort).
    }
    \label{fig:cell_type}
\end{figure*}

\section{BMEC Dataset}


Our BMEC dataset was built based on the data acquired at the Second Hospital of Jilin University, Changchun, China. 
It contains 5,666 erythroid cell images which are annotated into four fine-grained types by three hematologists.
The shape differences between the types are subtle, which makes the classification on BMEC more challenging than existing WBC datasets or RBC datasets with abnormal erythroid cells.

\begin{table}[h]
\begin{center}
\begin{minipage}{210pt}
\caption{The characteristics of different erythroid cells summarized in \cite{gregory2001bigger}. }\label{tab:chara}%

\begin{tabular}{@{}llll@{}}
\toprule
Cells  & Percent\footnotemark[1]  & Diameter & Shape\\
\midrule
Pro   & 0.5\%-4\% & 15um-20um & elliptical, protrusions \\
Bas   & 1\%-5\%   & 10um-18um & round, protrusions      \\
Pol   & 12\%-20\% & 8um-15um  & round, rough edges      \\
Ort   & 6\%-12\%  & 7um-10um  & round, smooth edges     \\
\botrule
\end{tabular}
\footnotetext[1]{The Percent column indicates the percentage of the corresponding cell type in all types of bone marrow cells, including both WBC and erythroid.}
\end{minipage}
\end{center}
\end{table}

\begin{table}[h]
\begin{center}
\begin{minipage}{\textwidth}
\caption{The statistics of the BMEC dataset in terms of distributions of different cells.}\label{tab:bmec}
\setlength{\tabcolsep}{2pt}
\begin{tabular*}{\textwidth}{@{\extracolsep{\fill}}lcccccc@{\extracolsep{\fill}}}
\toprule%
Dataset & Quantity & Percent & Pro\footnotemark[1]  & Bas\footnotemark[1] & Pol\footnotemark[1] & Ort\footnotemark[1] \\
\midrule
Training Set.   & 3,865& 68.2\%  & 146(2.6\%) & 428(7.6\%) & 1,702(30.0\%) & 1,589(28.0\%)\\
Validation Set  & 655  & 11.6\%  & 28(0.5\%)  & 69(1.2\%)  & 312(5.5\%)    & 246(4.4\%)\\
Test Set        & 1,146& 20.2\%  & 48(0.8\%)  & 115(2.0\%) & 526(9.3\%)    & 457(8.1\%)\\
Total           & 5,666& 100\%   & 222(3.9\%) & 612(10.8\%)& 2,540(44.8\%) & 2,292(40.5\%)\\
\botrule
\end{tabular*}
\footnotetext[1]{proerythroblast cells (Pro), basophilic erythroblast cells (Bas), polychromatophilic erythroblast cells (Pol), orthochromic erythrobla cells (Ort).}

\end{minipage}
\end{center}
\end{table}

\textbf{Smears Imaging and Cell Extraction.}
The original data contains 239 sets of bone marrow erythroid cell smears collected from 128 patients using the BEION V4.90 system between 2019 and 2021.
To meet the requirements of clinical evaluation and diagnosis, each smear contains more than 200 blood cells.
In the collection process, the hematologist first stained all smears using the Richter stain method.
Then, the observation of an Olympus BX43 microscope at 100X magnification was captured into images of 2592 $\times$ 1944 resolution.

To extract individual erythroid cells from the smear images, two experienced hematologists worked together to select a  ROI (region of interest) for each cell and crop the selections into cell images (Figure \ref{fig:cell_type}).
When extracting the cells, the hematologists mainly follow two principles: 1) not introduce too much background and noise; 2) preserve the whole cell shape to the maximum extent.
When one hematologist completed the cropping, the other hematologist examined the results to ensure the quality of the extracted images, e.g., no other cells rather except the central erythroid cell appear in the image.
In the end, 5,666 erythroid cell images are extracted from the blood smears.
Each cell image is resized to 224$\times$224 resolution.

\textbf{Data Annotation.}
Based on the growth and maturation, the erythroid cells are annotated into four types (Figure \ref{fig:cell_type} right):
proerythroblast (Pro), basophilic erythroblast (Bas), polychromatophilic erythroblast (Pol), and orthochromic erythrobla (Ort).
Note that we only focus on the erythroid cells which have normal shapes and do not consider the abnormal erythroid cells with elongated shapes such as in erythrocytesIDB \cite{gonzalez2014red}.
\rui{For reference, in Table \ref{tab:chara}, we show the characteristics (e.g., percentage, size and shape) summarized in \cite{gregory2001bigger}, for the four types of erythroid cells.}

As the shape differences of the collected erythroid cells are indeed very subtle, to reduce the subjectiveness in annotation, we invited three hematologists and designed a rigorous annotation process to obtain more consistent labels.
First, two hematologists independently annotated all the cell images without any interference.
The annotation results of the two hematologists were compared and the cell images with different assigned labels were re-annotated by the two hematologists for the second pass.  
If the re-annotation results were still different, the third hematologist would provide another annotation on these images, and the final cell category would be determined by voting the annotations from the three hematologists.

\textbf{Statistics.}
As shown in Table \ref{tab:bmec}, we divide the BMEC dataset into a training set, a validation set and a test set with an approximate ratio of 7:1:2.
It can be observed that for each set, the percentage of the four cell types follows the relative cell distribution mentioned in Table \ref{tab:chara}, e.g., the number of Pol is about four times of Bas.

In the following, we summarize the key characteristics that discriminate our BMEC dataset from other blood cell datasets:

\begin{itemize}
\item \textbf{The first fine-grained erythroid cell dataset.}
BMEC dataset is the first dataset that contains four fine-grained types of normal erythroid cells: Pro,  Bas, Pol, and Ort. It is more challenging than existing WBC or erythroid cell datasets since the shape differences between the types are subtle.
\item  \textbf{Largest number of erythroid cells}. As shown in Table \ref{tab:datasets},
BMEC is currently the largest dataset for erythroid cells, which is expected to enable more deep learning research on erythroid cells.
\item \textbf{Professional and rigorous annotation}. The images in the BMEC dataset were professionally annotated by three hematologists following a rigorous labeling process, which reduces the subjectiveness and ensures the quality of the annotation.
\item \textbf{Free public access}. We make the BMEC dataset freely available for all communities, with the hope that it can inspire more follow-ups research on erythroid cells.
\end{itemize}

\begin{figure*}[t]
    \includegraphics[width=1.0\linewidth]{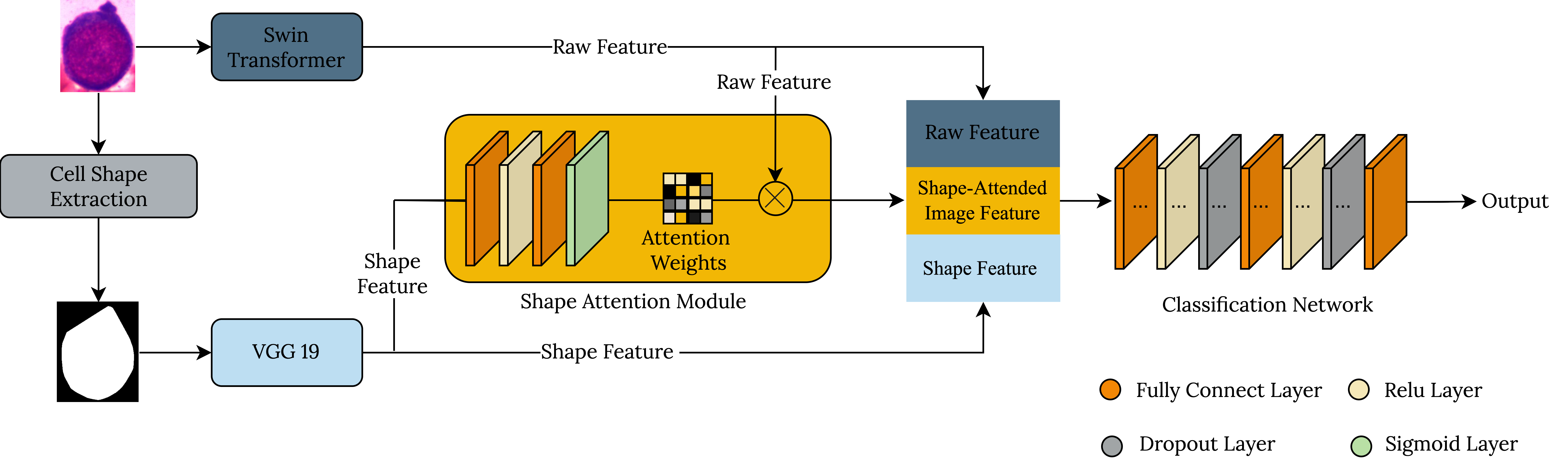}
    \caption{The overview of our approach. The novel shape attention module calculates a shape-attended image feature using the raw and shape features ($\otimes$ represents the broadcasted element-wise matrix multiplication operation). The fused shape-aware feature is used for the final fine-grained classification.  
    }
    \label{fig:overview}
\end{figure*}

\section{Method}


\subsection{Overview}


Our pipeline consists of four modules as shown in Figure \ref{fig:overview}: a cell shape extraction module, a dual-branch network, a shape attention module and a cell classification network.
The shape extraction module transforms the RGB erythroid cell image into the binary shape mask. The RGB image and shape mask are separately passed to the dual-branch network to generate the raw and shape features. The shape attention module calculates the shape attention weights and generates a shape-attended image feature from the raw image feature.
Then, the raw image feature, shape-attended image feature and the shape feature are aggregated to a fused shape-aware feature, which is passed to the lightweight classification network for prediction.

\begin{algorithm}[t]
\caption{The cell segmentation pipeline following \cite{tavakoli2021new}}\label{algo1}
\begin{algorithmic}[1]
\State Input RGB erythroid cell image 

\State Convert the input RGB image to CMYK color space 
\label{ code:fram:extract }
\State Calculate $KM = (K\_component) - (M\_component)$
\label{code:fram:trainbase}
\State Convert  the RGB input image to HLS color space and calculate $MS = Min(M\_component, S\_component)$
\label{code:fram:classify}
\State Calculate $V  = MS - KM$
\label{code:fram:select}
\State Exploit Otsu's thresholding algorithm \cite{NobuyukiOtsu1979ATS} to segment the cell and obtain the cell nucleus $N$
\State Calculate the convex hull of the cell to acquire the shape mask $S$ based on $N$
\State Calculate the cell cytoplasm $C = S - N$ 
\State Output the convex hull $S$, nucleus $N$, cytoplasm $C$ as the binary segmentation masks
\end{algorithmic}
\label{alg:Framwork}
\end{algorithm}


\subsection{Cell Shape Extraction}

\begin{figure*}[t]
    \centering
    \includegraphics[width=1.0\linewidth]{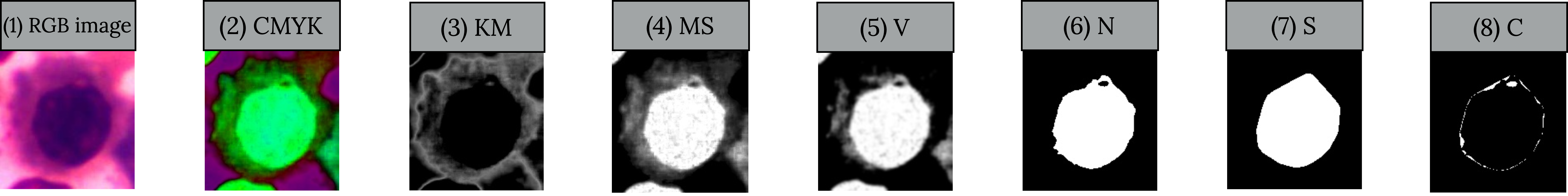}
    \caption{ The results obtained by applying different steps of Algorithm 1. 
    }
    \label{fig:interm}
\end{figure*}

\begin{figure*}[t]
    \centering
    \includegraphics[width=1.0\linewidth]{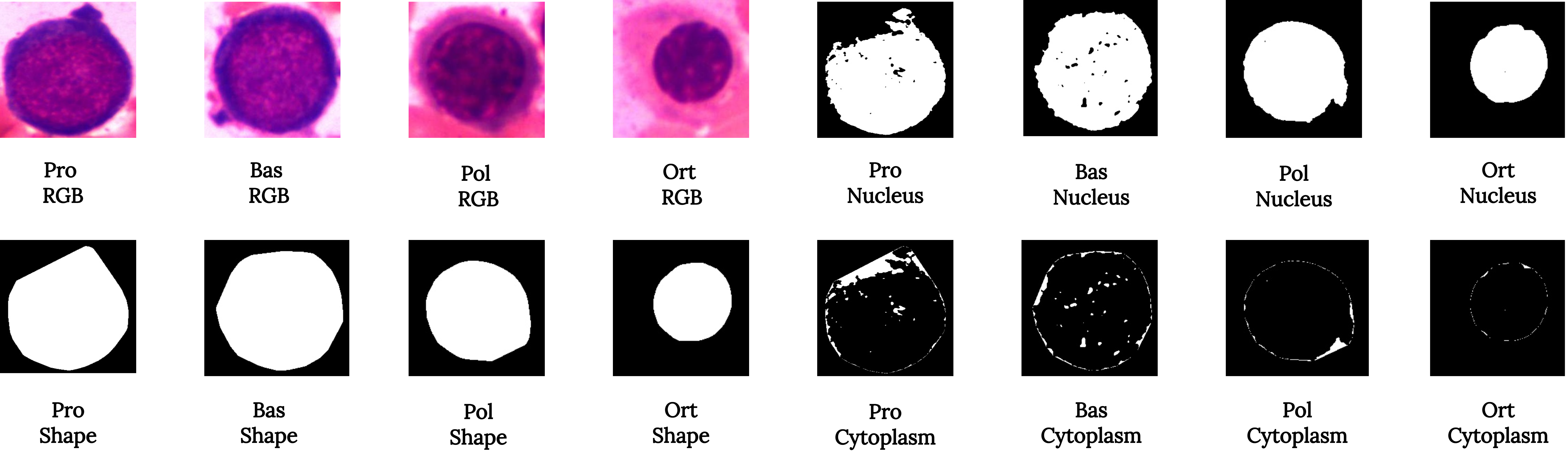}
    \caption{ This figure shows the RGB images and other segmentation results (Nucleus mask, Shape mask, and Cytoplasm mask) of four types of erythroid cells.
    } 
    \label{fig:seg_results}
\end{figure*}

From Figure \ref{fig:cell_type} and Table \ref{tab:chara}, it can be observed the shape information is crucial to distinguish different fine-grained types of erythroid cells, especially when the appearances of the cells are similar.
Therefore, we explicitly extract the cell shape information as a binary shape mask and treat it as a prior knowledge (or inductive bias) for the classification model. 
We employ the cell image segmentation method proposed in \cite{tavakoli2021new} to extract the shape mask for each BMEC cell image.
The main steps of cell shape extraction are shown in Algorithm \ref{alg:Framwork}, while Figure \ref{fig:interm} shows the corresponding intermediate and segmentation results.
Three types of binary mask images are obtained: the convex hull, nucleus and cytoplasm.
Figure \ref{fig:seg_results} shows the image segmentation results of different types of erythroid cells.

Unlike \cite{tavakoli2021generalizability} which utilized the nucleus and ROC (convex hull - nucleus) images as the shape images, we choose the convex hull image as our shape image.
The main reason is \cite{tavakoli2021generalizability} mainly focuses on the WBCs whose nucleuses are in apparently different shapes.
However, for erythroid cells, their nucleuses are more intact and look quite similar comparing to WBCs (see Figure \ref{fig:seg_results}).
Also, the cytoplasm (i.e., the ROC in \cite{tavakoli2021generalizability}) images of the erythroid cells are noisy and not suitable to represent the cell shapes.
In contrast, convex hull contains the key shape characteristics and it is more robust to noise comparing to the nucleus.
We pass the binary image of convex hull to our network and also compute the shape attention to enhance the original image feature.

\subsection{Dual-Branch Network}
To explicitly integrate the shape information into the deep learning framework, we design a dual-branch network which encodes the RGB image and the binary shape mask (convex hull image) in two separate branches.

As shown in Figure \ref{fig:overview}, the top branch network takes the RGB image as input and extracts the raw feature using Swin-Transformer \cite{liu2021swin}, the state-of-the-art backbone model for image feature extraction.
Although the RGB image branch can learn the pixel intensity variations between different cells, it does not focus on the cell's shape or structure characteristics.
Therefore, we introduce a separate bottom branch network which employs a VGG19 \cite{he2016deep} model to explicitly compute the shape feature from the binary shape mask image obtained from the cell shape extraction module.
The raw image feature and the shape feature from two branches are integrated in the shape attention module (Section \ref{sec:shape_att}) and fused in the later stage.
In Section \ref{sec:ablation}, we show the shape branch is an effective complement to the RGB branch for fine-grained cell classification.
In addition, our dual-branch network is a general design such that the backbone models in each branch can be replaced by other deep image feature extraction models.
In current our experiments, we find using the combination of Swin-Transformer and VGG19 for the two branches can achieve the best performance (Table \ref{tab:backbone_top} and \ref{tab:backbone_bottom}).

\subsection{Shape Attention}
\label{sec:shape_att}


To further integrate the shape information to the extracted image feature, we propose a novel shape attention module which takes the raw image feature and shape feature as input and outputs a shape-attended image feature (see Figure \ref{fig:overview} middle).
First, the shape attention weights are computed from the binary shape mask image following the similar attention mechanism used in SENet \cite{JieHu2018SqueezeandExcitationN}.
Then, the raw image feature is multiplied by the shape attention weights to get the shape-attended feature.
The mathematical formulation is as follows:

\begin{equation}
\label{eq:fr}
    F_R = Swin-Transformer(X_R),
\end{equation}
\begin{equation}
\label{eq:fs}
    F_S = VGG19(X_S),
\end{equation}
\begin{equation}
\label{eq:wa}
    W_{SA} = Sigmoid(W_2(ReLU(W_1(F_S)))),
\end{equation}
\begin{equation}
\label{eq:fa}
    F_{SA} = W_{SA}\otimes F_R.
\end{equation}
\\



\noindent Here, $X_R \in R^{224\times 224 \times 3}$ and $X_S \in R^{224\times 224 \times 1}$ are the input RGB image and binary shape mask, respectively;
$F_R$ and $F_S$ are the corresponding features extracted using Swin-Transformer \cite{liu2021swin} and VGG19 \cite{simonyan2014very}.
As shown in Equation \ref{eq:wa} and Figure \ref{fig:overview}, the shape attention weights $W_{SA}$ are calculated by passing $F_S$ through two fully connected layers ($W_1$ and $W_2$) which are followed by a ReLU and a Sigmoid layer, respectively.
Finally, the shape-attended feature $F_{SA}$ is obtained by performing element-wise multiplication of the shape attention weights $W_{SA}$ and the raw image feature $F_R$.
In summary, the shape attention module is a lightweight network that can efficiently computes the correlation among the shape feature and enhances the raw image feature with self-learned cell shape information.
Moreover, the shape attention module can be combined with any backbone models to integrate the shape information to the image feature.


\subsection{Cell Classification Network}
The last module of our pipeline is a conventional classification network which consists of two sets of fully connected, ReLu and Dropout layers (Figure \ref{fig:overview} right).
The raw image feature $F_R$, shape feature $F_S$ and shape-attended image feature $F_{SA}$ are concatenated to form an fused cell feature $F_C = \{F_R, F_{SA}, F_S\}$ for final classification. 
With the shape feature and shape attention explicitly considered, the shape-aware network can achieve superior performance comparing to only using the RGB feature for the fine-grained erythroid cell classification (see Table \ref{tab:ablation}).

\section{Experiments}

	
    




    

In this section, we first provide the implementation details of our method.
Then, we qualitatively evaluate the performance of different backbones in our dual-branch network and show the combination of Swin-Transformer (for RGB image) and VGG19 (for shape mask) can work best on our fine-grained BMEC dataset.
To further verify the effectiveness and generalizability of our shape-aware network, we compare the results on two WBC datasets and show our method can consistently achieve the best accuracy on different datasets.
In the end, we perform an ablation study to show the contribution of the shape attention module for improving the classification accuracy.
The code and BMEC dataset can be found on \url{https://github.com/wangye8899/BMEC}.

\subsection{Implementation Details}

The network architecture in Figure \ref{fig:overview} is implemented with the Timm \cite{rw2019timm} and PyTorch \cite{paszke2017automatic}. Both branch networks (Swin-Transformer, VGG19) are initialized with ImageNet pre-trained weights.
Furthermore, the shape attention module and cell classification network are initialized with Kaiming initialization \cite{KaimingHe2015DelvingDI}.
We adopt the SGD optimizer with a momentum of 0.9, weight decay of 2e-5
and a cosine lr schedule with the warm-up strategy \cite{loshchilov2016sgdr}. 
We set the warm-up learning rate to 1e-4, the maximum lr to 1e-3, and the minimum lr to 1e-4. 
Four data augmentation techniques are used: random crop, random horizontal flip with probability of 0.5, random vertical flip with probability of 0.5, and color jitter with factor of 0.4.
The network is trained for 300 epochs with batch size of 32 on \rv{a cloud-based server with a 4.90GHz CPU and} three NVIDIA RTX 3090 24G GPUs and the total training takes 3 hours.
\rv{During inference, the computation time is instantaneous which is similar to other deep learning based classification methods.}

\subsection{Classification Results on BMEC}
Our dual-branch network allows flexible choices of different backbone models for the cell classification.
To quantitatively evaluate the performance of different backbones for the two branches, we conducted classification experiments with various classic and SOTA image feature extraction models on our BMEC dataset and compared the results in Table \ref{tab:backbone_top} and \ref{tab:backbone_bottom}.
Since the RGB image in the top branch contains more color variations than the binary shape mask in the bottom branch, it is expected that a more complex backbone model is needed for the top branch, while a relative simple model may be sufficient for the bottom branch.
Therefore, in Table \ref{tab:backbone_top}, we fixed the bottom branch with a VGG19 which is a classic CNN based model and tested different backbones for the top branch.
It is shown the SOTA Swin-Transformer model can produce the best accuracy and F1 score comparing to other models such as ResNet \cite{he2016deep}, DenseNet \cite{huang2017densely} or ResNest \cite{HangZhang2020ResNeStSN} for extracting features from the RGB images.
Similarly, we fixed the top branch network as Swin-Transformer and evaluated different models for the bottom branch.
The results in Table \ref{tab:backbone_bottom} shows the VGG19 model is most suitable for the bottom branch, while one possible reason is that other models may underfit or overfit the binary shape mask images.
Based on above experiments, we finally choose Swin-Transformer and VGG19 for the top and bottom branches, and our final model achieved 81.12\% accuracy on BMEC.
Note that the backbones can always be easily evaluated and replaced when newer and more powerful image models are proposed.

\begin{table}[t]
\begin{center}
\begin{minipage}{\textwidth}
\caption{Comparison of different top branch backbone networks for classification on BMEC. The bottom branch network is fixed as VGG19.}\label{tab:backbone_top}
\setlength{\tabcolsep}{3pt}
\begin{tabular*}{\textwidth}{@{\extracolsep{\fill}}lcccccc@{\extracolsep{\fill}}}
\toprule%
Top & Bottom & Inputs& Acc & Pre& Rec& F1  \\
\midrule
ResNet18 \cite{he2016deep}         & \multirow{5}{*}{VGG19 \cite{simonyan2014very}} & RGB+Shape & 79.51\%  & 74.27\%   & 73.53\% & 72.42\%\\
ResNet50 \cite{he2016deep}         &                        & RGB+Shape & 79.58\%  & 74.33\%   & 73.67\% & 72.51\% \\
DenseNet121 \cite{huang2017densely}      &                        & RGB+Shape & 79.65\%  & 73.42\%   & 73.15\% & 72.20\% \\
ResNest \cite{HangZhang2020ResNeStSN}          &                        & RGB+Shape & 80.12\%  & \textbf{76.47\%}   & 74.64\% & 73.04\% \\
Swin-T \cite{liu2021swin}\footnotemark[1] &                        & RGB+Shape & \textbf{81.12\%}  & 76.28\%   & \textbf{75.73\%} & \textbf{74.00\%} \\

\botrule
\end{tabular*}
\footnotetext[1]{Swin-Transformer (Swin-T)}

\end{minipage}
\end{center}
\end{table}

\begin{table}[t]
\begin{center}
\begin{minipage}{\textwidth}
\caption{Comparison of different bottom branch backbone networks for classification on BMEC. The top branch network is fixed as Swin-Transformer.}\label{tab:backbone_bottom}
\begin{tabular*}{\textwidth}{@{\extracolsep{\fill}}lcccccc@{\extracolsep{\fill}}}
\toprule%
Top & Bottom & Inputs& Acc & Pre& Rec& F1  \\
\midrule
\multirow{4}{*}{Swin-T \cite{liu2021swin}\footnotemark[1]} 
                                  & ResNet18 \cite{he2016deep}         & RGB+Shape & 80.62\%  & 78.68\%   & 74.93\% & 73.23\% \\ 
                                  & ResNet50 \cite{he2016deep}         & RGB+Shape & 80.45\%  & 75.44\%   & \textbf{76.21\%} & 72.68\% \\ 
                                  & Swin-T \cite{liu2021swin}\footnotemark[1] & RGB+Shape & 80.36\%         & \textbf{77.45\%}          & 75.81\%        & 73.33\%        \\ 
                                  & VGG19 \cite{simonyan2014very}            & RGB+Shape & \textbf{81.12\%}  & 76.28\%   & 75.73\% & \textbf{74.00\%} \\

\botrule
\end{tabular*}
\footnotetext[1]{Swin-Transformer (Swin-T)}

\end{minipage}
\end{center}
\end{table}

\subsection{Classification Results on WBCs Datasets}

To further demonstrate the effectiveness of our shape-aware network, we conducted classification experiments on two other public WBCs datasets, LISC \cite{rezatofighi2011automatic} and Raabin-WBC \cite{kouzehkanan2021raabin}.
We retrained our model using their training data and compared the testing results with the reported numbers in \cite{rezatofighi2011automatic} and \cite{kouzehkanan2021raabin}.
As shown in Table \ref{tab:lisc} and \ref{tab:rabbin}, we achieved the best classification accuracy on both datasets, showing the generalizability of our method.
In addition, the high accuracy (over 98.5\%) on both WBC datasets indicates the classification of WBCs is relatively easier than the erythroid cells since the shape differences of WBCs are more prominent.
We hope our BMEC dataset can inspire and facilitate more research on fine-grained classification and other learning tasks for erythroid cells.

\begin{table}[t]
\begin{center}
\begin{minipage}{\textwidth}
\caption{Comparison of different methods for classification on LISC dataset \cite{rezatofighi2011automatic}. Due to the data imbalance problem in LISC, other methods directly discarded the cell categories with a low number of images. In contrast, we trained and tested using all categories and still obtained the best accuracy.}\label{tab:lisc}
\setlength{\tabcolsep}{3pt}
\begin{tabular*}{\textwidth}{@{\extracolsep{\fill}}lcccccc@{\extracolsep{\fill}}}
\toprule%
Study& Method& Category Number & Accuracy  \\
\midrule
Rezatofighi et al. (2011) \cite{rezatofighi2011automatic} & SVM                               & 5               & 96.00\%  \\ 
Jung et al. (2019) \cite{jung2019w}                      & CNN                   & 5               & 97.00\%  \\

Baydilli et al. (2020) \cite{baydilli2020classification}              & Capsule Networks        & 5               & 96.86\%  \\ 
Harshanand et al. (2020)  \cite{harshanand2020comprehensive}              & CNN         & 5               & 97.64\%  \\ 
Tavakoli et al. (2021) \cite{tavakoli2021new}                        & Segmentation                      & 5               & 92.21\%  \\ 
Muhammad et al. (2021) \cite{khan2021automated}            & Feature Selection and ELM        & 4              & 96.60\%  \\ 

Ours                      & DB and SA\footnotemark[1] & 6              & \textbf{98.51\%}    \\ 

\botrule
\end{tabular*}
\footnotetext[1]{Dual Branch and Shape Attention (DB and SA)}

\end{minipage}
\end{center}
\end{table}

\begin{table}[t]
\begin{center}
\begin{minipage}{\textwidth}
\caption{Comparison of different methods for classification on Raabin-WBC dataset \cite{kouzehkanan2021raabin}.}\label{tab:rabbin}
\begin{tabular*}{\textwidth}{@{\extracolsep{\fill}}lcccccc@{\extracolsep{\fill}}}
\toprule%
Study& Method& Category Number & Accuracy \\ 
\midrule
Kouzeh  et al. (2021) \cite{kouzehkanan2021raabin}            & VGG16        & 6              & 98.09\%  \\ 
Kouzeh et al. (2021) \cite{kouzehkanan2021raabin}            & MnasNet1        & 6              & 98.59\%  \\ 

Kouzeh et al. (2021) \cite{kouzehkanan2021raabin}            & DenseNet121        & 6              & 98.87\%  \\ 
Kouzeh et al. (2021) \cite{kouzehkanan2021raabin}            & ShuffleNet-V2        & 6              & 99.03\%  \\
Tavakoli et al. (2021) \cite{tavakoli2021new}            & Segmentation        & 6              & 94.65\%  \\ 

Ours                     & DB and SA\footnotemark[1] & 6               & \textbf{99.17\%}    \\ 

\botrule
\end{tabular*}
\footnotetext[1]{Dual Branch and Shape Attention (DB and SA)}

\end{minipage}
\end{center}
\end{table}

\begin{table}[t]
\begin{center}
\begin{minipage}{\textwidth}
\caption{Ablation study on the shape information using our BMEC dataset.}\label{tab:ablation}
\setlength{\tabcolsep}{3pt}
\begin{tabular*}{\textwidth}{@{\extracolsep{\fill}}lcccccc@{\extracolsep{\fill}}}
\toprule%
Top & Bottom & Inputs& Acc & Pre& Rec& F1  \\
\midrule
\multirow{2}{*}{ResNet18 \cite{he2016deep}}         & ---                 & RGB       & 78.79\% & 71.91\% & 73.09\%&
                                  71.77\%
\\ 
                                  & VGG19 \cite{simonyan2014very}               & RGB+Shape & 79.51\% & 74.27\%&
                                  73.54\%& 72.42\%    \\
                                  \hline
                                  
\multirow{2}{*}{ResNet50 \cite{he2016deep}}         & ---                 & RGB       & 79.31\%   & 72.16\%          
                                  & 73.18\%   & 71.85\%    \\ 
                                  & VGG19 \cite{simonyan2014very}               & RGB+Shape & 79.58\%    & 74.33\%          
                                  & 73.67\% & 72.51\%   \\ \hline
                                  
\multirow{2}{*}{DenseNet121 \cite{huang2017densely}}      & ---                 & RGB       & 79.30\%    & 71.45\%           
                                  & 72.23\%     & 70.91\%     \\ 
                                  & VGG19 \cite{simonyan2014very}               & RGB+Shape & 79.65\%   & 73.42\%
                                  & 73.15\% & 72.20\%    \\ \hline
                                  
\multirow{2}{*}{ResNest \cite{HangZhang2020ResNeStSN}}          & ---           & RGB       & 80.09\%    & 74.49\%          
                                  & 73.95\%      &72.87\%    \\ 
                                  & VGG19 \cite{simonyan2014very}               & RGB+Shape & 80.12\%    & \textbf{76.47\%}
                                  & 74.64\% & 73.04\%   \\ \hline
                                  
\multirow{2}{*}{Swin-T \cite{liu2021swin}\footnotemark[1]} & ---                 & RGB       & 80.62\%    & 75.81\%                                         &74.79\%     & 72.85\%    \\  
                                  & VGG19 \cite{simonyan2014very}               & RGB+Shape & \textbf{81.12\%}    & 76.28\%
                                  &\textbf{75.73\%} & \textbf{74.00\%}    \\

\botrule
\end{tabular*}
\footnotetext[1]{Swin-Transformer (Swin-T)}

\end{minipage}
\end{center}
\end{table}

\subsection{Ablation Study on Shape information}
\label{sec:ablation}




To verify the effectiveness of the shape information, we performed an ablation study which compared our full dual-branch network with the RGB-only single branch network (see Table \ref{tab:ablation}).
Specifically, we disabled the bottom branch and only used the raw image feature for classification.
In this case, the network becomes a conventional RGB image based classification model.
When two branches are enabled, both the shape feature extracted from the binary shape mask and shape-attended image feature computed from the shape attention module are fused to the raw image feature for classification.
For the single branch network, different backbone models are tested on the BMEC RGB images.
For our dual-branch network, the VGG19 is used in each configuration so that we can compare the performance improvement for each case.
From the results in Table \ref{tab:ablation}, the dual-branch network can always outperform the single-branch, confirming the effectiveness of the shape information for improving the fine-grained cell classification.

\section{Conclusion, Limitation and Future Work}

We introduce BMEC, the first and largest professionally annotated image dataset for fine-grained classification of erythroid cells.
The BMEC is public available for all communities and we hope it can encourage more deep learning research on erythroid cells.
We propose a novel shape-aware network which explicitly utilizes the shape information to improve the classification accuracy in a dual-branch deep learning framework.
\rv{We conduct extensive evaluations of different backbones and perform experiments on BMEC and two other WBC datasets.
The results show the superiority and generalizability of our shape-aware dual-branch network.
Ablation studies also verify the effectiveness of the shape information for the fine-grained cell classification.
}

Although our method has achieved superior performance on cell classification, it still has some limitations which can inspire more future work.
Our current cell shape extraction still involves several image processing steps which may be affected by the image noise and staining quality.
It will be interesting to automatically learn the cell shape segmentation in the network and then apply the learned shape mask in our dual-branch network.
Meanwhile, our classification works on the cropped cell images which are prepared in the BMEC dataset.
Developing a cell detection and classification network which can directly work on the full-field cell images is also a promising future work.
Due to the privacy issue, we cannot release the full-field cell images used in our BMEC, but it is still possible to train a cell classification network using the cropped BMEC cell images.

\section*{Declarations}

\begin{itemize}
\item Funding: This research is supported by the National Natural Science Foundation of China (Grants No. 61772227, 61972174, 61972175, 62202199), 
Science and Technology Development Foundation of Jilin Province (No. 20180201045GX, 20200201300JC, 20200401083GX, 20200201163JC), 
the Jilin Development and Reform Commission Fund (No. 2020C020-2).
\item Conflict of interest/Competing interests: The authors declare no conflict of interest.
\item Availability of data and materials: The code and dataset have been published on github (\url{https://github.com/wangye8899/BMEC}).
\item Code availability: \url{https://github.com/wangye8899/BMEC}
\item Authors' contributions: Conceptualization, Y.W. and R.M.; methodology, Y.W., R.M. and Y.Z.; validation, X.M., Y.X. and X.W.; formal analysis, Y.W. and R.M.; investigation, H.C.; resources, Y.Z.; data curation, H.C. and X.M.; writing---original draft preparation, Y.W.; writing---review and editing, R.M.; visualization, Y.W. and R.M.; supervision, Y.Z.; project administration, Y.Z.; All authors have read and agreed to the published version of the manuscript.
\end{itemize}









\bibliography{sn-article}


\end{document}